\pdfoutput=1

\documentclass[11pt]{article}

\usepackage[]{acl}

\usepackage{times}
\usepackage{latexsym}

\usepackage[T1]{fontenc}

\usepackage[utf8]{inputenc}

\usepackage{microtype}

\usepackage{enumerate}
\usepackage[inline]{enumitem}
\usepackage{amsmath, amssymb}

\usepackage{subcaption}
\usepackage{graphicx}
\usepackage{booktabs}
\usepackage{diagbox}
\usepackage{rotating}
\usepackage{makecell, tabularx, multirow}
\usepackage{longtable}

\newcommand\blfootnote[1]{%
  \begingroup
  \renewcommand\thefootnote{}\footnote{#1}%
  \addtocounter{footnote}{-1}%
  \endgroup
}

\title{High-Resource Methodological Bias in Low-Resource Investigations}

\author{Maartje ter Hoeve$^*$ \\
  University of Amsterdam \\
  \texttt{m.a.terhoeve@uva.nl} \\\And
  David Grangier \\
  Apple MLR \\
  \texttt{grangier@apple.com} \\\And
  Natalie Schluter \\
  Apple MLR \\
  \texttt{natschluter@apple.com} \\}

\begin{document}
\maketitle

\blfootnote{*Work done while interning at Apple MLR.}

\begin{abstract}
The central bottleneck for low-resource NLP is typically regarded to be the quantity of accessible data, overlooking the contribution of data quality. This is particularly seen in the development and evaluation of low-resource systems via down sampling of high-resource language data. In this work we investigate the validity of this approach, and we specifically focus on two well-known NLP tasks for our empirical investigations: POS-tagging and machine translation. We show that down sampling from a high-resource language results in datasets with different properties than the low-resource datasets, impacting the model performance for both POS-tagging and machine translation. Based on these results we conclude that naive down sampling of datasets results in a biased view of how well these systems work in a low-resource scenario. 
\end{abstract}


\section{Introduction}
\label{sec:introduction}

The field of natural language processing (NLP) has experienced substantial progress over the last few years, with the introduction of neural sequence-to-sequence models~\cite[e.g.,][]{kalchbrenner-blunsom-2013-recurrent, vaswani2017attention} and large, pre-trained transformer based language models~\cite[e.g.,][]{devlin-etal-2019-bert, brown2020language}. Despite their impressive performance, these models require a lot of training resources, which are not always available. Approaches specifically targeted towards low-resource scenarios try to address this issue~\cite[e.g.,][]{agic-etal-2016-multilingual, plank-agic-2018-distant, zhu-etal-2019-importance, bai-etal-2021-cross}. Resource scarcity manifests itself in various ways, such as a lack of compute power~\cite[e.g.,][]{hedderich-etal-2020-transfer} or a lack of (labeled) training data~\cite[e.g.,][]{adelani-etal-2021-masakhaner}. In this work we focus on the latter. 

Whether or when a scenario or language should be considered as ``low-resourced'' has been topic of debate~\cite[e.g.,][]{bird-2022-local}. In this work, we add to this discussion by highlighting that many low-resource approaches are grounded in high-resource scenarios, as has also been noted previously ~\cite[e.g.,][]{kann2020weakly}. This is problematic from a cultural or sociolinguistic perspective~\cite[e.g.,][]{hamalainen2021endangered, bird-2022-local}, as well as from a methodological perspective~\cite[e.g.,][]{kann2020weakly}. Although both perspectives are arguably intertwined, we mostly focus on the latter in this work.

For example, a popular approach to develop and evaluate low-resource systems is to down sample uniformly from a high-resource language to simulate a low-resource scenario~\cite[e.g.,][]{fadaee-etal-2017-data, araabi-monz-2020-optimizing, chronopoulou-etal-2020-reusing, ding-etal-2020-daga, kumar-etal-2021-machine}. The motivations for this setup are often justifiable, for example if used to investigate the effect of the dataset size, or because low-resource data is hard to obtain. However, we do believe that there are two potential issues with this down sampling approach, that should be carefully considered.

Firstly, a large dataset is potentially much richer in content than a small dataset, for example in terms of the number of domains that are covered, the number of different styles, etc. That is, the vocabulary size of a large dataset is expected to be larger than the vocabulary size of a small dataset. 
This would affect the vocabulary of the down sample, causing a mismatch between the down sampled dataset and the real low-resource scenario, potentially affecting the scores on the task at hand.

Secondly, we need to consider how datasets are constructed. On the one hand, there are examples of small, low-resource datasets, that are carefully constructed for a specific task~\cite[e.g.,][]{ter2020conversations, adelani-etal-2021-masakhaner, ter2022summarization}. Obtaining high quality data points is costly, and thus, once the dataset size increases, a different trade-off between quality and cost needs to be made~\cite[e.g.,][]{caswell-etal-2020-language, luccioni-viviano-2021-whats}. As~\citet{kreutzer-etal-2022-quality} point out, this trade-off can also affect the quality of low-resource languages in large multilingual datasets. For these large datasets, the quality and usefulness come from the size of the dataset, but not necessarily from the quality of each individual data point~\cite{kreutzer-etal-2022-quality}. When simulating a low-resource language by taking a uniform sample from a high-resource language, this quality-cost trade-off might result in a biased sample, as the sample might actually be of lower quality than can be expected in a truly low-resource scenario. This also links our work to approaches like active learning~\cite{cohn1996active} and curriculum learning~\cite{bengio2009curriculum}, that focus on the most helpful data points at any time during training.

More theoretically, we can summarize these points by taking a look at the estimation error that is optimized during training, typically in the form of a cross-entropy loss:

\begin{multline}
    \mathcal{L}(\theta; D) = 
    - \frac{1}{|D|} \underset{y \in D}{\sum} P_{D} (y) \log P_{M}(y | \theta),
\end{multline}

in which $D$ refers to the data, $M$ to the model, $y$ to the prediction and $\theta$ to the model parameters. Uniformly down sampling to the same size as the simulated low-resource dataset deals with the $1 / |D|$ term, but it does not account for the fact that $D$ itself is different in the low- and high-resource setting. This mismatch is also referred to as the \textit{proxy fallacy}~\cite{agic-vulic-2019-jw300}.

In this work we investigate the effect of simulating a low-resource scenario by taking a uniform down sample from a high-resource setting in the context of two well-known NLP tasks: part-of-speech (POS)-tagging and machine translation (MT). We empirically find evidence for both issues raised above: 
\begin{enumerate*}[label=(\roman*)]
    \item down sampling from a high-resource scenario increases the richness of the vocabulary of the sample, and 
    \item the quality of the high-resource dataset is sometimes lower than the low-resource variant.
\end{enumerate*}
As such, our work serves as a reminder to be careful when simulating low-resource scenarios by uniformly down sampling from a high-resource dataset.


\section{Related work}
\label{sec:related_work}

In this section we first discuss the definition of `low-resource' (Section~\ref{sec:related_work_def}). We then continue with a discussion of low-resource approaches in the NLP literature (Section~\ref{sec:related_work_nlp}). 
We end with a discussion on different training strategies (Section~\ref{sec:related_work_learning}).

\subsection{On the Definition of `Low-Resource'}
\label{sec:related_work_def}

Despite the amount of work on low-resource languages, or low-resource scenarios, it is hard to find a definition of when a scenario, or even a language, counts as low- or high-resource. It seems questionable to call a language low-resourced if it is spoken by millions of people who communicate in oral and/or written form in that language~\cite[e.g.,][]{hamalainen2021endangered, bird-2022-local}. In this work we do not explicitly define when a scenario is considered to be low-resource, but instead we use a more relative approach. That is, we will compare languages with different amounts of written data available, which is mainly indicated by the availability of the datasets that we use. In that sense we follow the implicit definition as used in previous work~\cite[e.g.,][]{zhu-etal-2019-importance, hedderich-etal-2021-survey}.

\subsection{Low-Resource Approaches in NLP}
\label{sec:related_work_nlp}

With the recent surge of work on NLP systems that require a lot of resources~\cite[e.g.,][]{devlin-etal-2019-bert, brown2020language, chowdhery-etal-2022-palm}, the question of designing systems that also work in a low-resource scenario has received a lot of attention. We refer to~\citet{hedderich-etal-2021-survey} for a recent survey. Although there are many examples of approaches that ground themselves in a `truly' low-resource scenario~\cite[e.g.,][]{plank-etal-2016-multilingual, kann2020weakly, adelani-etal-2021-masakhaner} (but see the discussion in Section~\ref{sec:related_work_def} above), there are also many examples of approaches where assumptions are made that are more plausible in a higher resource scenario~\cite[e.g.,][]{li-etal-2012-wiki, gu-etal-2018-meta, ding-etal-2020-daga, liu-etal-2021-mulda}. 
For example,~\citet{kann2020weakly} investigate the POS-tagging performance when no additional resources, like manually created dictionaries, are available, and they find that performance drops substantially. As mentioned in Section~\ref{sec:introduction}, our work focuses on the validity of the common approach to simulate a low-resource scenario by randomly down sampling from a higher resource dataset~\cite[e.g.,][]{gu-etal-2018-meta, chronopoulou-etal-2020-reusing, dehouck-gomez-rodriguez-2020-data, kumar-etal-2021-machine, park-etal-2021-unsupervised, zhang-etal-2021-two}.

\subsection{Different Learning Strategies}
\label{sec:related_work_learning}

Different learning strategies have been proposed to optimally make use of available data. Curriculum learning (CL)~\cite{bengio2009curriculum} is motivated by the idea that humans learn best when following certain curricula. For example, one effective curriculum is to learn new things in increasing order of difficulty. CL aims at finding similar curricula for artificial model training, by finding meaningful orders in which to present data to a model, such that the model learns more effectively. Some studies report improved results when using CL~\cite[][]{xu2020curriculum, chang-etal-2021-order, zhang2021reducing}, whereas for other studies CL does not seem to help yet~\cite[e.g.,][]{liu2019roberta, rao-vijjini-etal-2021-analyzing}.

Active learning (AL)~\cite{cohn1996active} is a related learning strategy, in which a model actively selects the data that it can be most effectively trained on at different points during the training process, for example based on its uncertainty for certain data points. Because of this property, AL has often been used as an effective way to decide which data points to label in an unlabeled dataset~\cite[e.g.,][]{reichart-etal-2008-multi, xu-etal-2018-using, ein-dor-etal-2020-active, chaudhary-etal-2021-reducing}.

\section{Empirical Investigation}
\label{sec:empirical}

In this section we empirically investigate down sampling from a high- to a low-resource scenario on two well-known NLP tasks: POS-tagging and machine translation. Both tasks are also popular low-resource tasks~\cite[e.g.,][]{hedderich-etal-2021-survey, haddow-etal-2022-survey} for which down sampling strategies have been used~\cite[e.g.,][]{irvine-callison-burch-2014-hallucinating, ding-etal-2020-daga, kann2020weakly, araabi-monz-2020-optimizing}, making them suitable for our investigation. Moreover, POS-tagging is especially suitable as the task is relatively quick and straight forward, giving us a good starting point. We found down sampling approaches to be especially prominent in the MT literature~\cite[e.g.,][]{irvine-callison-burch-2014-hallucinating, fadaee-etal-2017-data, ma-etal-2019-key, araabi-monz-2020-optimizing, kumar-etal-2021-machine, xu-etal-2021-improving-multilingual}, making it a natural task for our investigation. Our work serves as a good starting point to investigate other tasks in future work. For each task we investigate the effect of down sampling on the dataset statistics, and on the modeling performance for the task.

We emphasize that our goal is to get a general understanding of the effect of simulating a low-resource scenario by randomly down sampling from a high-resource scenario. Therefore, we also keep our investigation general. That is, we use default versions of state-of-the-art models for both tasks, instead of versions that are fully optimized to get the highest possible scores. We also explicitly do not dissect individual papers in which down sampling is used. This is not the goal of this work, and we believe that there can be good reasons to use down sampling, as discussed in Section~\ref{sec:introduction}. Instead, we aim to provide useful insights that can be taken into consideration in future work.

\subsection{POS-tagging}
 
Briefly, POS-tagging is the task of assigning grammatical parts of speech, such as nouns, verbs, etc., to tokens in the input text. We use the Universal Dependencies (UD) dataset (see ~\citet{de-marneffe-etal-2021-universal} for a recent description) for our experiments (Section~\ref{sec:ud_explanation}). In the first part of our POS-tagging investigation we show that down sampling indeed increases the richness of the sample in terms of vocabulary size (Section~\ref{sec:pos_downsample_ds_stats}). Next, we show that an increased vocabulary size positively affects the performance in POS-tagging tasks (Section~\ref{sec:pos_downsample_modeling}).

\subsubsection{Data Description}
\label{sec:ud_explanation}
The Universal Dependencies project\footnote{Website: \url{https://universaldependencies.org/}, Github: \url{https://github.com/UniversalDependencies}.} consists of treebanks for over a hundred languages~\cite{de-marneffe-etal-2021-universal}, with varying amounts of resources. Languages are labeled with morphosyntactic labels, such as dependency tags and POS-tags. We only make use of the POS-tags. 

\subsubsection{Effect of Down Sampling on Dataset Statistics}

In the first part of our investigation, we down sample datasets from several high-resource languages, until they have the same size as the lower resource language datasets in the UD. We determine size based on the number of tokens or sentences. A natural question to ask at this point is whether tokens in different languages can be equally compared from a typological point of view. Therefore, we start with a typological inspection of different languages in the UD collection. 

\paragraph{Typological considerations.} 
Languages differ from each other in their morphological complexity, for example in their morpheme per word ratios~\cite{baker2012taal}. Although subject to some debate, this can be described as the difference between analytic and synthetic languages.\footnote{There are also still other categories, like isolating languages. As we simply base ourselves on the morpheme per word ratios for our analysis, we leave these out for simplicity.} Analytic languages have a low morpheme per word ratio, as opposed to synthetic languages. Within the synthetic category, one can differentiate between agglutinative and fusional languages, depending on how well single morphemes can be distinguished.

To the best of our knowledge, there is no easily accessible, exhaustive list that categorizes the languages in the UD as either analytic or synthetic. We approach the categorization using two proxies. First, we use the \textit{inflectional synthesis of the verb} as reported by the WALS~\cite{wals-22}.\footnote{\url{https://wals.info/chapter/22}} This feature measures the number of inflectional categories per verb in different languages. To do so, it uses the ``most synthetic'' form of the verb. WALS defines $7$ categories, ranging from $0$-$1$ till $12$-$13$ categories per word.
We label all UD languages that are included in the WALS for this feature.
Second, if Wikipedia pages exist for the languages in the UD, and they give information about the language type, we use this as a proxy to label the corresponding UD languages.

Motivated by the idea that the language type might affect the tokenization quality, we compute the average ratio between the unique number of tokens and the total number of tokens for the labeled languages (Table~\ref{tab:typology_scores}). We only find a significant difference between the agglutinative and analytic languages ($t=-2.20, p=0.04$). Agglutinative languages have more unique tokens per total of tokens, so they could be harder to tokenize. However, as we will see next, even if we down sample from an analytic language like English, we end up with a larger vocabulary size in the majority of samples.

\begin{table}
\centering
\begin{tabular}{llrr}
\toprule
& \multicolumn{1}{l}{\textbf{Category}} & \multicolumn{1}{r}{\textbf{Count}} & \multicolumn{1}{r}{\textbf{Avg ratio}} \\
\midrule
\parbox[t]{2mm}{\multirow{4}{*}{\rotatebox[origin=c]{90}{\emph{WALS}}}}
        &  $0-1$ & $1$ & $0.12 \scriptscriptstyle{\pm 0.00}$  \\
        &  $2-3$  & $6$ & $0.12 \scriptscriptstyle{\pm 0.07}$  \\
        &  $4-5$ & $10$ & $0.09 \scriptscriptstyle{\pm 0.05}$ \\
        &  $6-7$ & $5$ & $0.18 \scriptscriptstyle{\pm 0.10}$ \\
     \midrule
\parbox[t]{2mm}{\multirow{3}{*}{\rotatebox[origin=c]{90}{\emph{Wiki}}}}
&  Analytic   & $9$  & $0.11 \scriptscriptstyle{\pm 0.03}$ \\
&  Agglutinative  & $22$ & $0.22 \scriptscriptstyle{\pm 0.15}$  \\
&  Fusional   & $4$ & $0.10 \scriptscriptstyle{\pm 0.05}$  \\ 
\bottomrule
\end{tabular}
\caption{Average ratio of vocabulary size per total number of tokens for different language types.}
\label{tab:typology_scores}
\end{table}

\paragraph{Investigation of data statistics.}
\label{sec:pos_downsample_ds_stats}
With these typological considerations in mind, we now proceed to investigating the effect of down sampling on the dataset statistics. The UD provides an excellent test bed for our inspection, as the datasets of the included languages are of different sizes. First, we filter them on a number of criteria: 

\begin{enumerate}[leftmargin=*,nosep]
\item We only include non-extinct languages;
\item We only include languages that have a POS-tagged dataset available on the UD Github page;
\item For some corpora, the tokens are not released but instead marked by an underscore. We filter these out;
\item Some languages have multiple corpora that are very similar, but somewhat differently tagged. Japanese is an example. We filter these corpora to avoid duplication.
\end{enumerate} 

\noindent Based on these selection criteria, we arrive at a total of $100$ languages. A full overview of all languages and corpora that we consider can be found in Appendix~\ref{sec:appendix_ud_languages}. We select the five highest resource languages in the UD: Czech, French, German, Icelandic, and Russian. We also include English, as it is often used to down sample from and still one of the higher resourced languages in the UD.

\begin{figure*}
\begin{subfigure}{\textwidth}
  \centering
  \includegraphics[width=\textwidth]{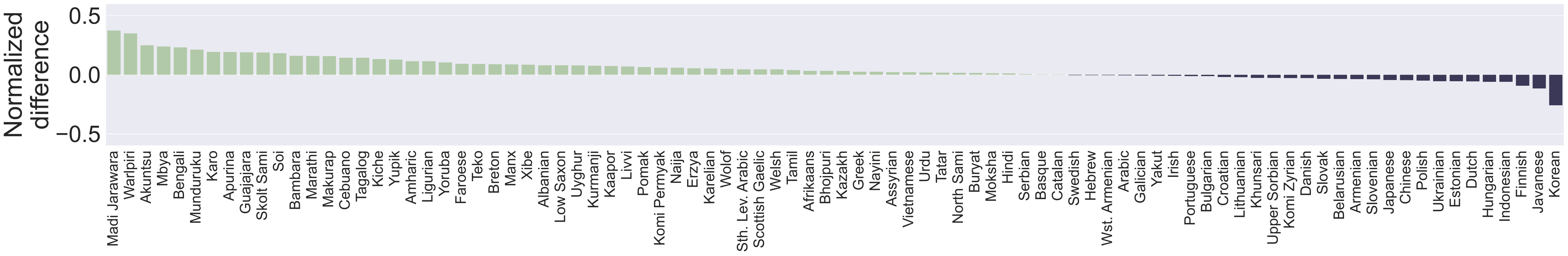}  
  \caption{Down sample English}
  \label{fig:ds_en}
\end{subfigure}
\begin{subfigure}{\textwidth}
  \centering
  \includegraphics[width=\textwidth]{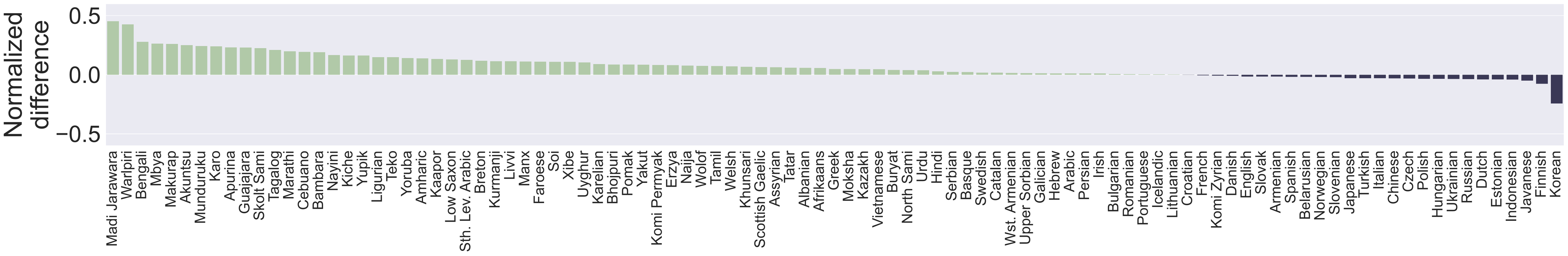}  
  \caption{Down sample German}
  \label{fig:ds_de}
\end{subfigure}
\begin{subfigure}{\textwidth}
  \centering
  \includegraphics[width=\textwidth]{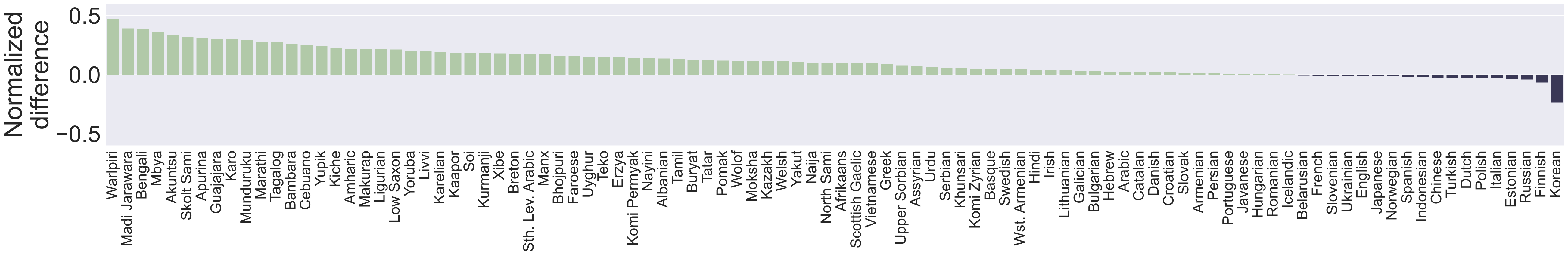}  
  \caption{Down sample Czech}
  \label{fig:ds_cz}
\end{subfigure}
\begin{subfigure}{\textwidth}
  \centering
  \includegraphics[width=\textwidth]{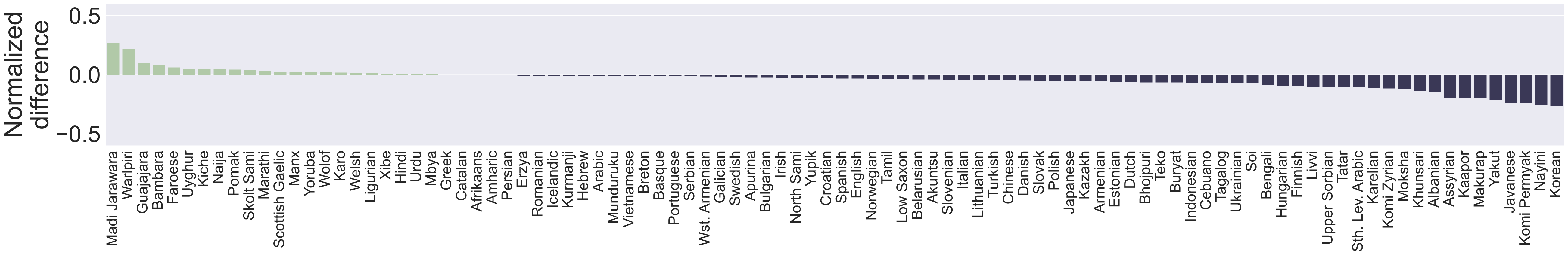}  
  \caption{Down sample French}
  \label{fig:ds_fr}
\end{subfigure}
\begin{subfigure}{\textwidth}
  \centering
  \includegraphics[width=\textwidth]{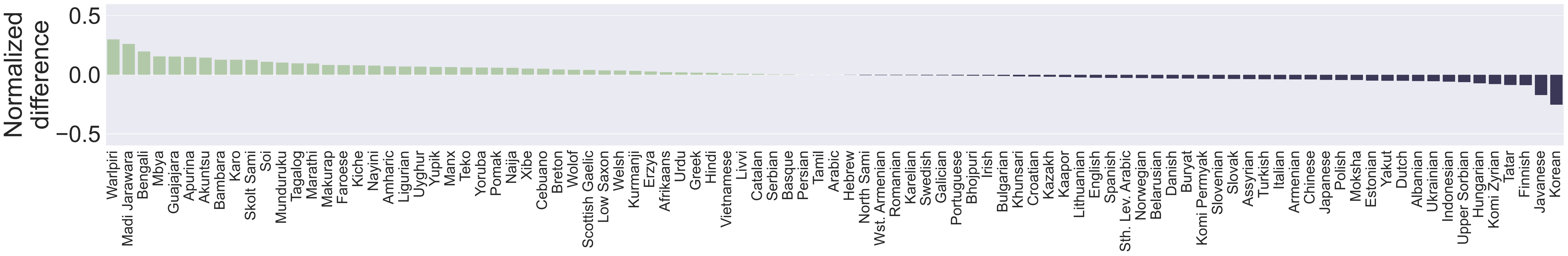}  
  \caption{Down sample Icelandic}
  \label{fig:ds_ic}
\end{subfigure}
\begin{subfigure}{\textwidth}
  \centering
  \includegraphics[width=\textwidth]{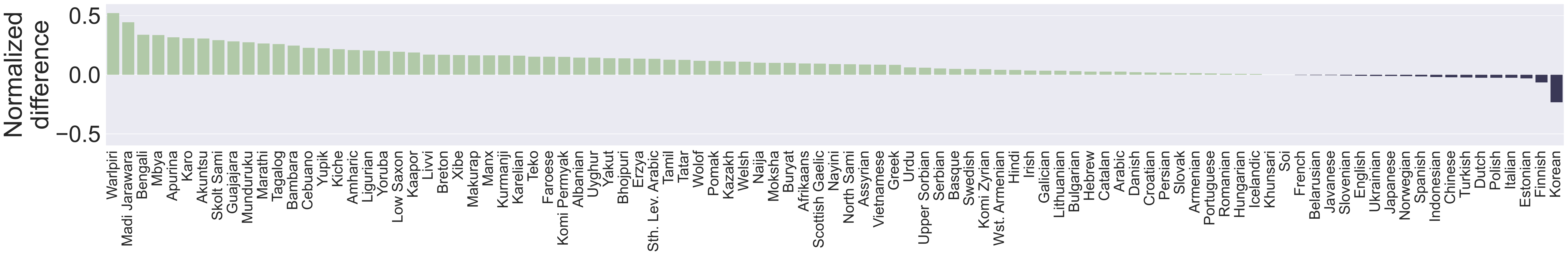}  
  \caption{Down sample Russian}
  \label{fig:ds_ru}
\end{subfigure}
\caption{Effect of down sampling on vocabulary size. Down sampling based on number of tokens. Each plot describes a different language that we are down sampling. The x-axis shows the language that we use as reference. The normalized difference in vocabulary size between the down sample and the original low-resource reference language is shown on the y-axis. A positive number indicates that the vocabulary size of the down sample is larger than the original low-resource dataset, whereas a negative number indicates the opposite. We find that down sampling indeed results in a larger vocabulary in the vast majority of cases.}
\label{fig:ds_effect_ud_stats_tok_based}
\end{figure*}

Next, we randomly down sample each of these high-resource languages to the size of the remaining lower resource languages. We compute size based on number of tokens and number of sentences. We report the results based on number of tokens in the main body of the paper. We want to know how down sampling affects the vocabulary size. Therefore, we compute the difference in vocabulary size between the down sampled dataset and its respective low-resource dataset. We normalize by the number of tokens in the low-resource dataset, to make a fair comparison. We plot the results of this analysis in Figure~\ref{fig:ds_effect_ud_stats_tok_based}. In this plot, a positive number indicates that the vocabulary size of the down sample is larger than the original low-resource dataset, whereas a negative number indicates the opposite. We find that down sampling indeed results in a larger vocabulary in the vast majority of cases. This is exactly in line with our intuition from Section~\ref{sec:introduction}. We find the same effect for down sampling based on number of sentences (Appendix~\ref{sec:appendix_pos_plots}, Figure~\ref{fig:ds_effect_ud_stats_sent_based}). For this setting we also compare how the total number of tokens in the down sampled datasets compare with the original low-resource datasets. We plot the results in Appendix~\ref{sec:appendix_pos_plots}, Figure~\ref{fig:ds_effect_ud_stats_sent_based_toks_per_sents}. We find that the down sampled corpora mostly contain more tokens than their originals.

\subsubsection{Effect of Down Sampling on Model Training}
\label{sec:pos_downsample_modeling}

Having shown that down sampling from a higher resource dataset often results in a larger vocabulary than in the original lower resource language, we now investigate the effect of vocabulary size on the modeling performance for POS-tagging. In line with most related work, we now fully focus on English as our high-resource language. We sample a number of smaller datasets from the English UD. Each of these samples has the same number of sentences, but they differ in vocabulary size. To achieve this, we use a greedy approach for the down sampling: we shuffle all sentences and greedily add sentences until we have the desired vocabulary size and the desired number of sentences.\footnote{We also experimented with implementing token based down sampling, instead of sentence based. However, we did not find a good trade-off where the vocabulary size increased, whereas the number of tokens stayed the same. We also experimented with different methods than greedy sampling, but this did not change our findings.} Like this, we construct training datasets of $1{,}000$ sentences each, for three vocabulary sizes: $1{,}000$, $2{,}000$ and $3{,}000$ tokens. We limit the validation sets to the same vocabulary as the training set, and use the original test set in order to be able to compare different settings equally. We sample each of these settings five times, for five different random seeds.  

Next, we use these sampled datasets to model the POS-tagging task, for which we use the standard POS-tagging setup from the FlairNLP library.\footnote{\url{https://github.com/flairNLP/flair/blob/master/resources/docs/TUTORIAL_7_TRAINING_A_MODEL.md}} We use FlairNLP's implementation of a sequence-to-sequence tagger, which defaults to a bidrectional RNN-CRF.\footnote{\url{https://github.com/flairNLP/flair/blob/master/flair/models/sequence_tagger_model.py}}
We compare three different word embedding types: \begin{enumerate*}[label=(\roman*)]
\item \textit{word2vec} embeddings~\cite{mikolov2013distributed} that we train from scratch on our training sets,
\item pre-trained Glove embeddings, and
\item pre-trained BERT embeddings.
\end{enumerate*} For the latter two we use the implentation from FlairNLP, for the \textit{word2vec} embeddings we use Gensim.\footnote{\url{https://github.com/RaRe-Technologies/gensim}} This setting is most realistic, as it is the only embedding type that is trained without access to another dataset or model.
However, sometimes low-resource work still makes use of these large pre-trained models, which is why we include them. Moreover, a model like English BERT has been shown to be surprisingly multilingual~\cite{pires-etal-2019-multilingual}. The results are given in Table~\ref{tab:pos_tagging_micro_f1}.\footnote{For the setting with a vocabulary size of $1{,}000$ we had to remove the results of one of the seeds, as it did not find enough sentences.} We also give additional micro F1-scores in Appendix~\ref{sec:appendix_pos_scores}, Table~\ref{tab:pos_tagging_macro_f1}. We find that the model scores increase when the vocabulary size increases.\footnote{We also find that the scores for a vocabulary size of $2{,}000$ and $3{,}000$ tokens are similar, although the average for $3{,}000$ is higher.} In line with our down sampling analysis in the previous section, we find that the total number of tokens also increases. Unsurprisingly, we find that pre-trained word embeddings substantially outperform our own \textit{word2vec} model. 
\\

\noindent Summarizing, in our POS-tagging investigation we found that down sampling from high-resource languages often results in a richer vocabulary size. We also found that a larger vocabulary size positively affects the scores on the POS-tagging task, in our settings for English. This is in line with the first issue that we raised in Section~\ref{sec:introduction}.  Of course, our experiments did not cover all possible settings that one can encounter in a low-resource scenario, and there are many follow up questions in the space of POS-tagging alone. For this work, we decide to take our results on the POS-tagging experiments as a first strong indication that one needs to be careful with naive down sampling, as we already find differences in the current, still limited, scenario. Encouraged by these findings, we now shift our focus to one more task that is often the focus of low-resource investigations: machine translation.

\begin{table*}
\centering
\begin{tabular}{rrrrrr}
\toprule
  & & & \multicolumn{3}{c}{\textbf{Macro F1}} \\
  \cmidrule(lr){4-6}
\multicolumn{1}{r}{\textbf{Vocab size}} & \multicolumn{1}{r}{\textbf{Nr Sents}} & \multicolumn{1}{r}{\textbf{Nr Toks}} & \multicolumn{1}{r}{\textbf{Word2Vec}} & \multicolumn{1}{r}{\textbf{Glove}} & \multicolumn{1}{r}{\textbf{BERT}} \\
\midrule
$1{,}000 $ & $1{,}000 $ & $7{,}235.75 \pm 174.485$ & $0.328 \pm 0.021$ & $0.743 \pm 0.005$ & $0.921 \pm 0.003$\\ 
$2{,}000 $ & $1{,}000 $ & $11{,}252.0 \pm 227.885$ & $0.350 \pm 0.024$ & $0.773 \pm 0.005$ & $0.937 \pm 0.003$\\ 
$3{,}000 $ & $1{,}000 $ & $14{,}867.2 \pm 292.534$ & $\textbf{0.360} \pm \textbf{0.006}$ & $\textbf{0.778} \pm \textbf{0.010}$ & $\textbf{0.940} \pm \textbf{0.005}$\\ 
\bottomrule
\end{tabular}
\caption{POS-tagging scores for different vocabulary sizes, while keeping the number of sentences equal. We report macro F1-scores for different word embeddings.}
\label{tab:pos_tagging_macro_f1}
\end{table*}


\subsection{Machine Translation}

Machine translation aims at translating text from a source to a target language. Machine learning systems address this task primarily by learning from bilingual documents with corresponding human translations~\cite{koehn2020neural}. These systems have shown substantial progress in recent years~\cite[e.g.,][]{barrault-etal-2019-findings, barrault-etal-2020-findings, akhbardeh-etal-2021-findings} and have been applied to a growing number of language pairs~\cite[e.g.,][]{platanios-etal-2018-contextual, costa-etal-2022-language}.
We make use of the WMT datasets (see~\citet{akhbardeh-etal-2021-findings} for a recent overview as well as Section~\ref{sec:mt_wmt_description}) for our experiments. We again divide our investigation into two parts. In the first part we show that down sampling again increases the richness of the sample in terms of vocabulary size (Section~\ref{sec:mt_downsampling_ds_stats}). 
Next, we show that low-resource and down sampled high-resource training datasets on the same task yield models with different accuracy: the down sampled dataset leads to a less accurate translation system than the original low-resource dataset
(Section~\ref{sec:mt_downsampling_model_training}).

\subsubsection{Data Description}
\label{sec:mt_wmt_description}
The WMT is a collection of datasets for machine translation belonging to the WMT shared tasks, which were first organized in $2006$~\cite{koehn-monz-2006-manual}. The first WMT collection consisted of three European language pairs: English-German, English-French and English-Spanish. Since then, the WMT has been expanded each year, with additional translation pairs for the original language pairs, and with additional data for new language pairs and new tasks~\cite{callison-burch-etal-2007-meta, callison-burch-etal-2008-meta, callison-burch-etal-2009-findings, callison-burch-etal-2010-findings, callison-burch-etal-2011-findings, callison-burch-etal-2012-findings, bojar-etal-2013-findings, bojar-etal-2014-findings, bojar-etal-2015-findings, bojar-etal-2016-findings, bojar-etal-2017-findings, bojar-etal-2018-findings, barrault-etal-2019-findings, barrault-etal-2020-findings, akhbardeh-etal-2021-findings}. An especially large jump in resources was made in $2017$. This expansion gives us a unique opportunity to test the effect of down sampling. In our investigation we treat the early versions of the WMT as low-resource setting, and later versions of the WMT as high-resource setting. We focus on the English-German translation pairs.

\subsubsection{Effect of Down Sampling on Dataset Statistics}
\label{sec:mt_downsampling_ds_stats}

To explore the effect of down sampling on the dataset statistics, we use the WMT $2014$ German-English dataset as our low-resource dataset (WMT14), and the 2018 version as our high-resource dataset (WMT18). We focus on the English-German translation task. We again apply two types of down sampling: sentence based and token based. For the sentence based down sampling we shuffle the WMT18 dataset, and sample the same number of sentences as in the WMT14 dataset. For the token based down sampling, we also shuffle the WMT18 dataset, but now we greedily add sentences until we reach the same number of tokens as in the WMT14 dataset. 

\begin{figure}[ht!]
\begin{subfigure}{\columnwidth}
  \centering
  \includegraphics[width=\columnwidth]{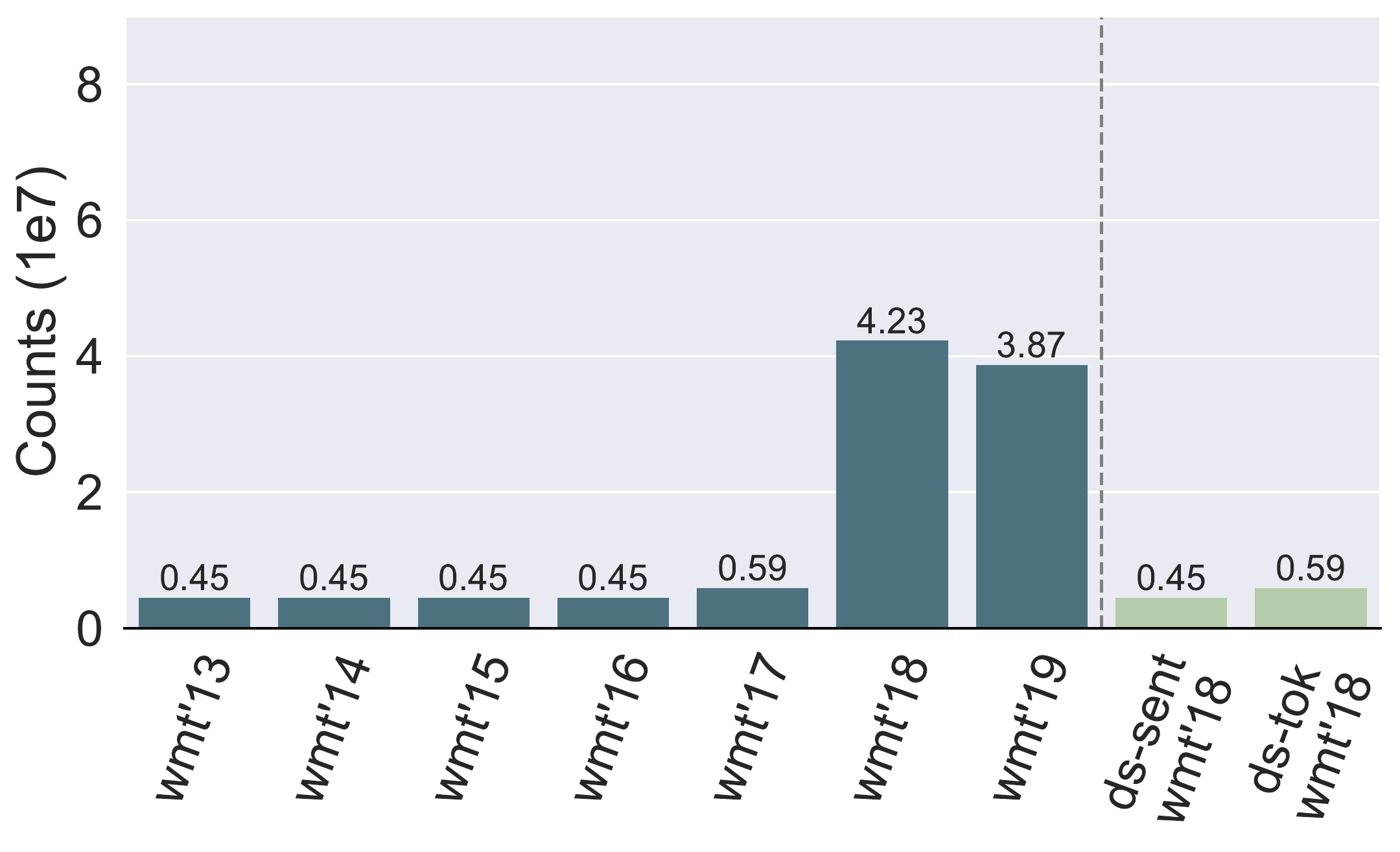}  
  \caption{Number of train sentences in WMT datasets.}
  \label{fig:train_wmt_sentences}
\end{subfigure}
\begin{subfigure}{\columnwidth}
  \centering
  \includegraphics[width=\columnwidth]{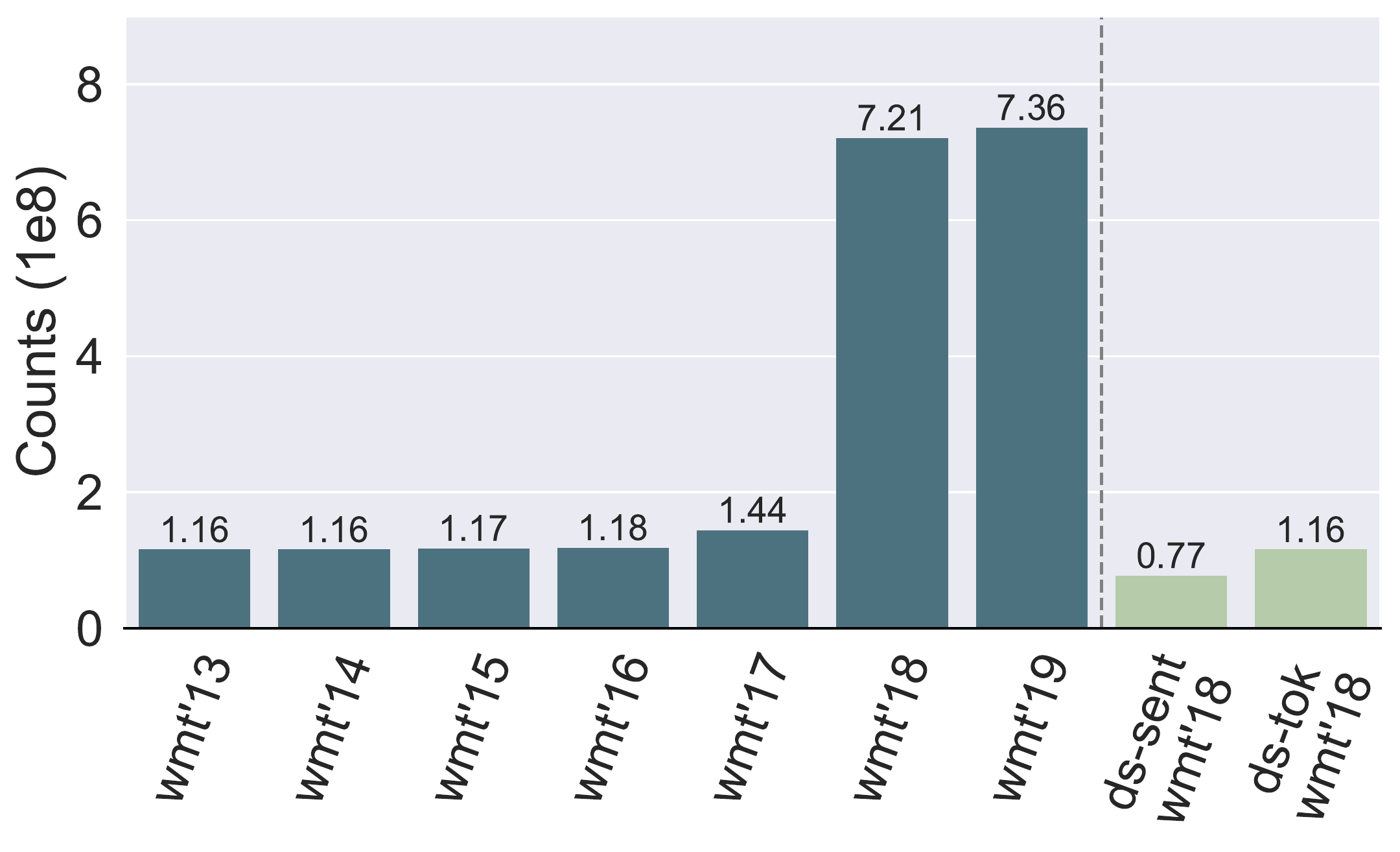}  
  \caption{Number of train tokens in WMT datasets.}
  \label{fig:train_wmt_tokens}
\end{subfigure}
\begin{subfigure}{\columnwidth}
  \centering
  \includegraphics[width=\columnwidth]{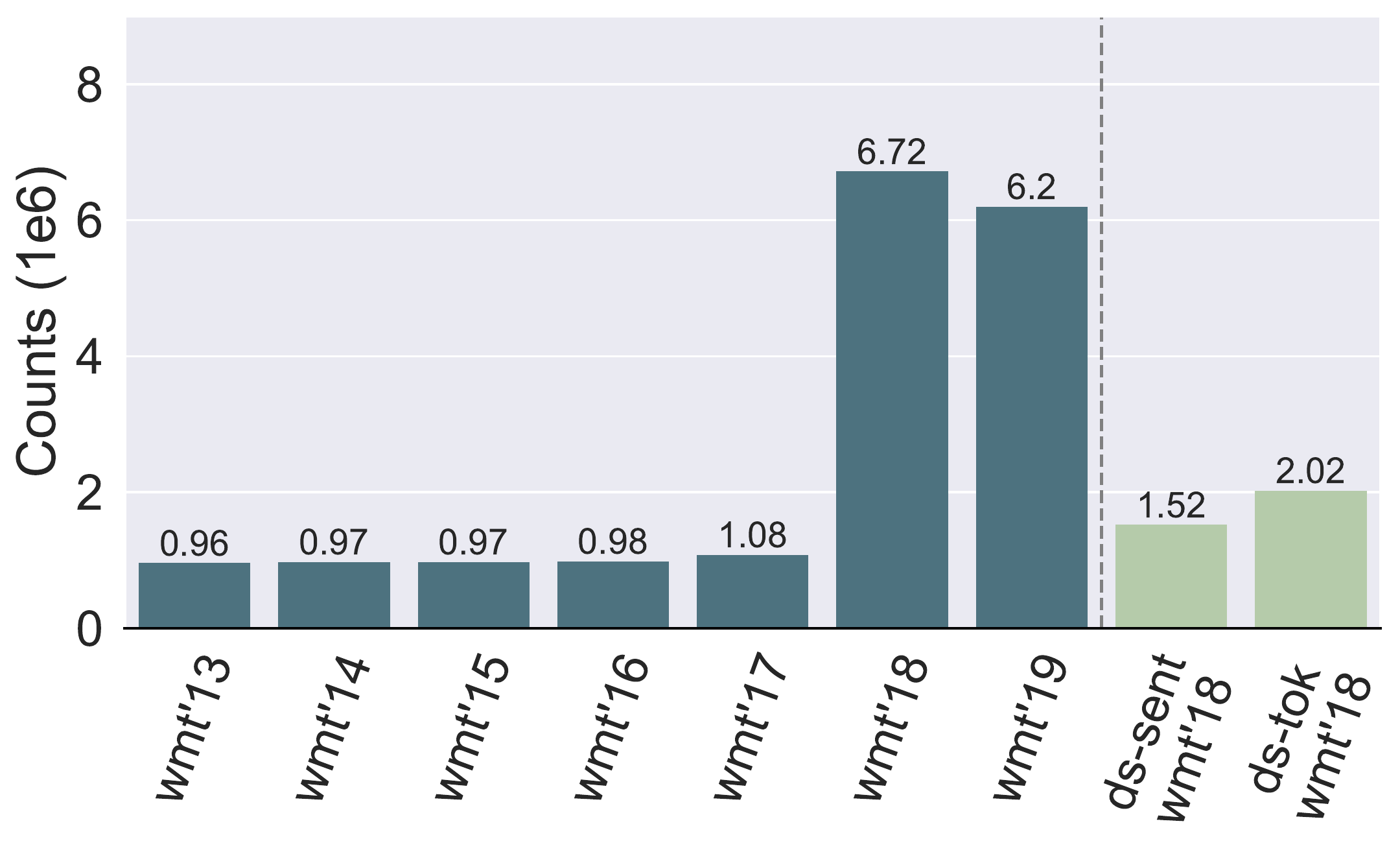}
  \caption{Vocabulary size of WMT datasets.}
  \label{fig:train_wmt_vocabb}
\end{subfigure}
\caption{Statistics of different WMT datasets (ds = down sampled, sent = sentence based, tok = token based).}
\label{fig:wmt_ds_stats}
\end{figure}

We plot the down sampling effect in Figure~\ref{fig:wmt_ds_stats}. These plots reflect the WMT train sets over different years. Even though we focus our investigations on WMT14 and WMT18, we plot all years from $2013$ till $2019$ for reference. The last two light green bars show our two down sampled datasets. If we down sample based on sentences (first light green bar right to the dotted line), we find that the number of tokens \textit{decreases}, whereas the vocabulary size \textit{increases}. If we down sample based on tokens, both the number of sentences and the vocabulary size increase.

We also qualitatively inspect the vocabulary distributions. In Appendix~\ref{sec:appendix_pos_plots}, Figure~\ref{fig:wmt_vocab_dists} we plot the $100$ most frequent words in each data set that we compare. We find that there are quite a few differences, especially in the second half of the plot.

\subsubsection{Effect of Down Sampling on Model Training}
\label{sec:mt_downsampling_model_training}

In this section we investigate the effect of down sampling on model training. 
To this end, we train and evaluate transformer sequence-to-sequence models~\cite{vaswani2017attention} in different data settings. We use the Flax transformer code\footnote{\url{https://github.com/google/flax/tree/main/examples/wmt}} for our implementation, and only adapt the data pipeline to be able to work with our down sampled datasets. We train these models on the standard WMT14 and WMT18 training sets, and on our two down sampled datasets (token and sentence based). We test on the WMT14 and WMT18 test sets (i.e., newstest data \cite{barrault-etal-2020-findings-first}). This leaves us with eight different settings in total. We report the scores in Table~\ref{tab:bleu_scores}. 

A few observations stand out. First, the models trained on down sampled versions of the WMT18 score lower on the WMT18 test set than the model trained on the original WMT18 dataset. This is as expected, if we assume that the additional WMT18 data would lead to better results. We also find that training on WMT14 and testing on WMT18 leads to higher scores than testing on the WMT14 test set. This is remarkable, but in line with earlier findings~\cite{edunov-etal-2018-understanding}. 
Finally, we observe that the models trained on the down sampled WMT18 datasets perform worse on the WMT14 test set than the models trained on the WMT14 dataset itself. This is the opposite finding from the POS-tagging experiments. 
For the MT experiments, having a richer vocabulary does not seem to help performance. We hypothesize that this can be explained by the quality of the WMT18 datasets, i.e., the second issue that we raised in Section~\ref{sec:introduction}. As shown in Figure~\ref{fig:wmt_ds_stats}, the amount of data increased heavily in $2017$, mostly driven by the inclusion of the Paracrawl data source~\cite{banon-etal-2020-paracrawl}. This data source is known to be noisy, and hence people have worked on filtering it~\cite[e.g.,][]{junczys-dowmunt-2018-dual, aulamo-etal-2020-opusfilter, zhang-etal-2020-parallel-corpus}.

\begin{figure}
  \centering
  \includegraphics[width=\columnwidth]{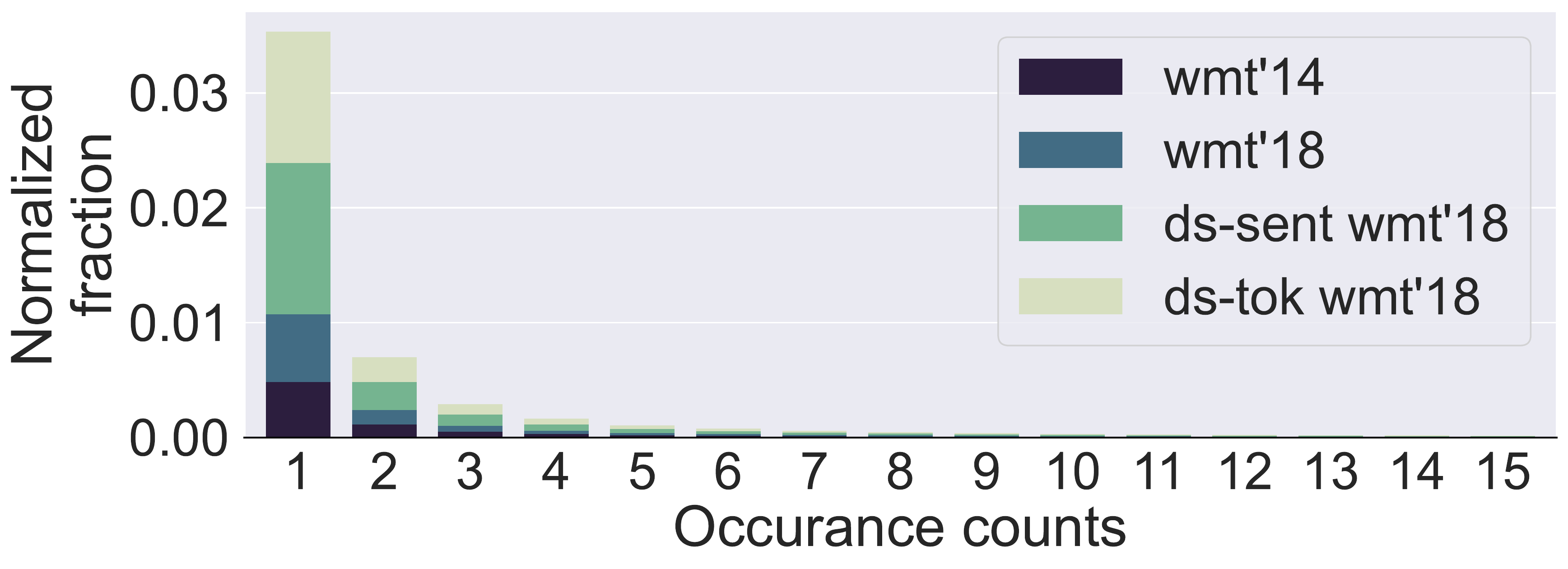}  
  \caption{How many words occur $N$ times in the different WMT datasets, normalized.}
  \label{fig:wmt_words_occur_n_times}
\end{figure}

To add to the investigation of the data quality, we count how many words occur $N$ times in the datasets, normalized by the total number of words in the datasets. The rationale behind this is that if a dataset contains many words that only occur once, this indicates that this dataset contains more noise (such as links) than datasets with fewer words that only occur once. We plot the results in Figure~\ref{fig:wmt_words_occur_n_times}. We find that WMT18 contains more words that only occur once than WMT14, an indicator that the average quality of the WMT18 dataset is indeed lower, negatively impacting our down sampled experiments.
\\

\noindent Summarizing, for our MT experiments we find that down sampling also increases the vocabulary size, in line with our hypothesis and with our findings for the POS-tagging experiments. We also found that the down sampled datasets did not increase the translation performance, which can be explained by the lower quality of the high resource data.

\begin{table}
  \small
  \centering
  \begin{tabular}{llrrrr}
  \toprule
  & &\multicolumn{4}{c}{\textbf{Train}} \\
  \cmidrule(lr){3-6}
    & \textbf{WMT} & \textbf{14} & \textbf{18} & \textbf{ds-sent-18} & \textbf{ds-tok-18} \\ 

  \midrule
  \multirowcell{-2}[-4ex]{\hspace*{-.6em}\turnbox{90}{\thead{\textbf{Test}}}} & \textbf{14} & $32.62$ & $32.12$ & $29.37$ & $30.23$ \\
  &\textbf{18} & $41.49$ & $39.70$ & $37.66$ & $38.20$ \\
  \bottomrule
  \end{tabular}
    \caption{BLEU scores MT experiments. Models are trained and tested on different (down sampled) train and test sets.} 
  \label{tab:bleu_scores}
\end{table}


\section{Discussion}
In our experiments we found evidence for both issues that we raised regarding simulating a low-resource scenario by taking a uniform down sample from a high-resource language. In this section we reflect on these findings. 
We hope that our work serves as additional evidence for the proxy fallacy of using high-resource methodologies for low-resource investigations. Being aware of this fallacy puts individual researchers and the field as a whole in a better position. Clearly, the best strategy is to use truly low-resource data whenever possible when conducting low-resource experiments. Fortunately, there are many examples of works that do this, or that only perform low-resource experiments on high-resource data for additional data points~\cite[e.g.,][]{kann2020weakly, kumar-etal-2021-machine, adelani-etal-2021-masakhaner}. 
There can still be good reasons why using truly low-resource data is not an option, for example because the type of data that is needed is just not available. In this case, we first want to echo~\citet{hedderich-etal-2020-transfer}, who show that by only labeling very few data points large improvements can already be made. We believe that we can also use recommendations from active learning and curriculum learning to choose which data points are best to label. We hope to experiment with this question in future work. If one is truly bound to simulating a low-resource scenario by using a high-resource language, one needs to be aware of the fallacies that we found in this work. The down sampled dataset is likely not a good reflection of the low-resource setting, which can result in scores that are either too high (because of the richness of the data) or rather too low (because the high-resource data may be of insufficient quality). 

\section{Conclusion}

In this work we investigated the validity of simulating a low-resource scenario by down sampling from a high-resource dataset. We argued that this process might be a poor proxy for a truly low-resource setting, for two reasons:     \begin{enumerate*}[label=(\roman*)]
        \item a high-resource dataset might be much richer in content than a low-resource dataset, and
        \item the high-resource dataset might be of lower quality than a low-resource dataset that was carefully crafted.
\end{enumerate*}
We empirically studied this on two well-known NLP tasks: POS-tagging and machine translation. Our investigation showed that uniform down sampling is indeed a poor proxy in these two scenarios, and we found evidence for both hypothesized reasons. As such, our work serves as a warning for work in low-resource domains. 
This work also serves as a starting point to formalize best practices to grow datasets, and to more reliable simulations of low- to high-resource settings. In future work, we plan to expand our analysis to more tasks and more languages. 

\section{Limitations}
Throughout this work we flagged some of the limitations of our approach. In this section we summarize these in more detail, to help future investigations.

\paragraph{The datasets.} In this work we concentrated on corpora from two data sources: the UD and the WMT. Although this is a good start and these datasets are a good fit for our investigation, we hope that future work investigates different corpora, to get an even better understanding of the effect of uniform down sampling. 

\paragraph{The tasks.} The same holds for the types of tasks that we chose. Although we believe POS-tagging and MT to be a good start, future work should investigate different tasks to be able to form a more general understanding.

\section{Ethical Statement}

In this work we developed an understanding for the effect of simulating a low-resource language by down sampling uniformly from a high-resource language. By pointing out biases that occur, we hope to have raised awareness for this issue, making follow-up work on low-resource languages more inclusive. However, there are around $7{,}000$ languages world-wide, of which we have only been able to cover a few.  

\bibliography{custom}
\bibliographystyle{acl_natbib}

\appendix


\section{Additional Plots POS-tagging Experiments}
\label{sec:appendix_pos_plots}

\begin{figure*}
\begin{subfigure}{\textwidth}
  \centering
  \includegraphics[width=\textwidth]{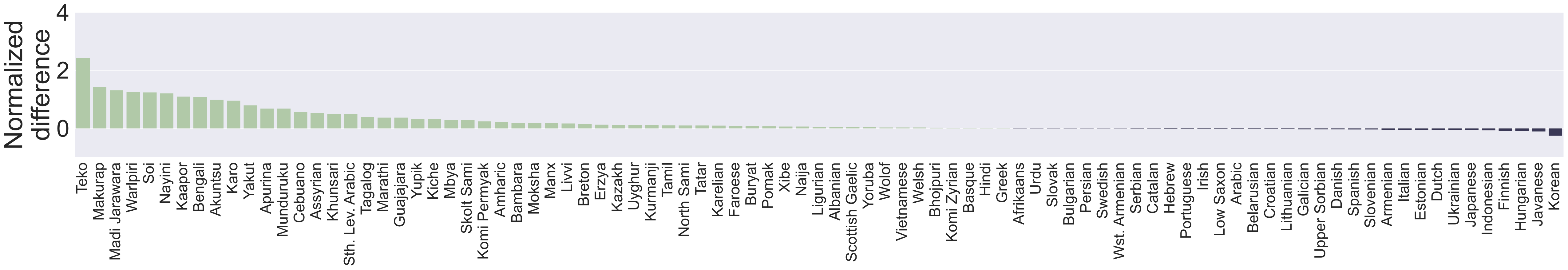}  
  \caption{Down sample English}
\label{fig:sent_ds_en}
\end{subfigure}
\begin{subfigure}{\textwidth}
  \centering
  \includegraphics[width=\textwidth]{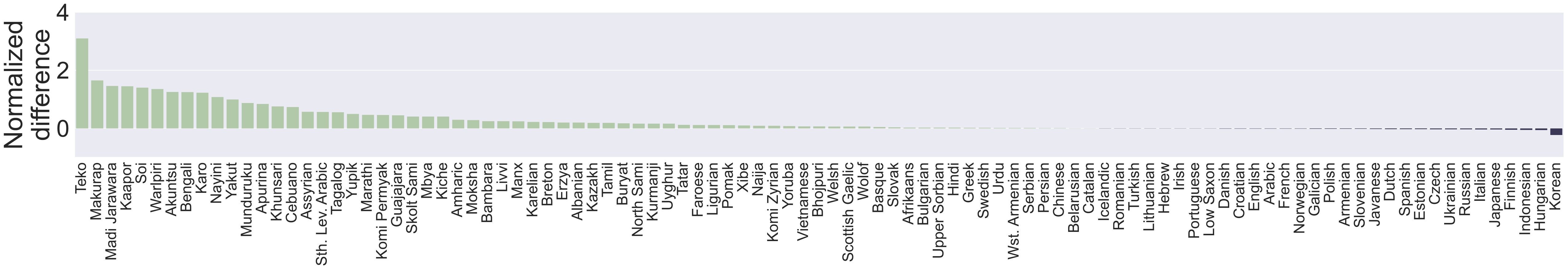}  
  \caption{Down sample German}
  \label{fig:sent_ds_de}
\end{subfigure}
\begin{subfigure}{\textwidth}
  \centering
  \includegraphics[width=\textwidth]{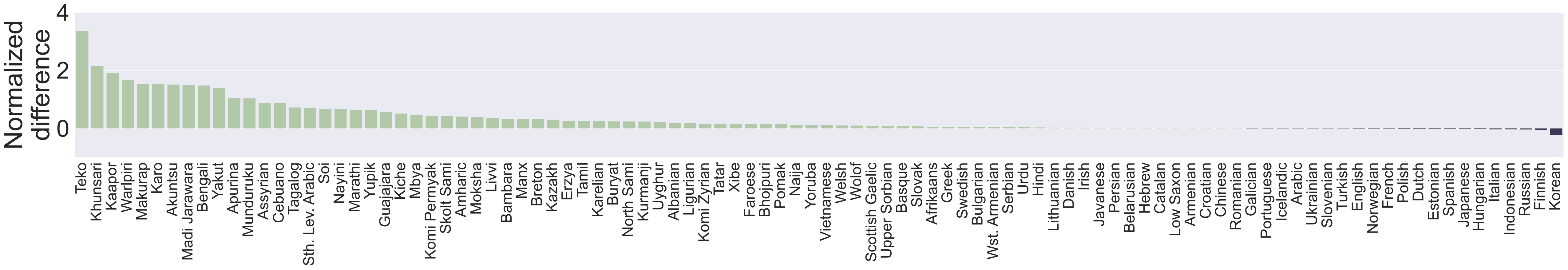}  
  \caption{Down sample Czech}
  \label{fig:sent_ds_cz}
\end{subfigure}
\begin{subfigure}{\textwidth}
  \centering
  \includegraphics[width=\textwidth]{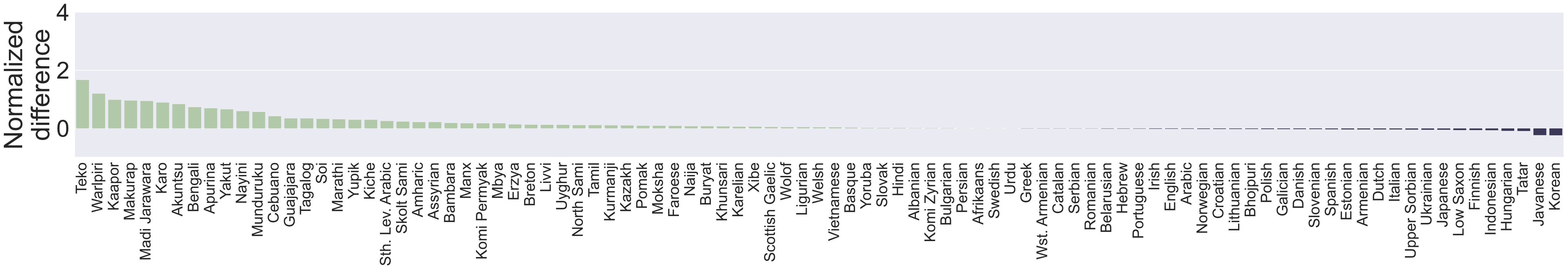}  
  \caption{Down sample French}
  \label{fig:sent_ds_fr}
\end{subfigure}
\begin{subfigure}{\textwidth}
  \centering
  \includegraphics[width=\textwidth]{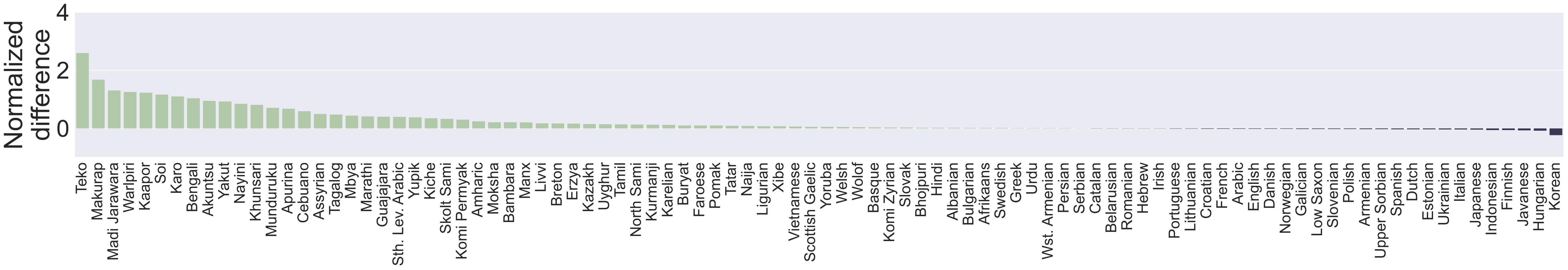}  
  \caption{Down sample Icelandic}
  \label{fig:sent_ds_ic}
\end{subfigure}
\begin{subfigure}{\textwidth}
  \centering
  \includegraphics[width=\textwidth]{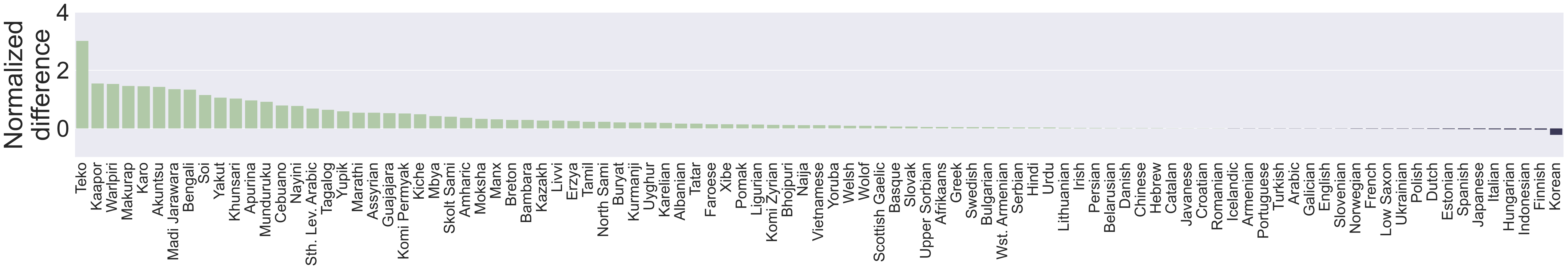}  
  \caption{Down sample Russian}
  \label{fig:sent_ds_ru}
\end{subfigure}
\caption{Effect of down sampling on vocabulary size. Down sampling based on number of sentences. Each plot describes a different language that we are down sampling. The x-axis describes the language that we use as reference.}
\label{fig:ds_effect_ud_stats_sent_based}
\end{figure*}

\begin{figure*}
\begin{subfigure}{\textwidth}
  \centering
  \includegraphics[width=\textwidth]{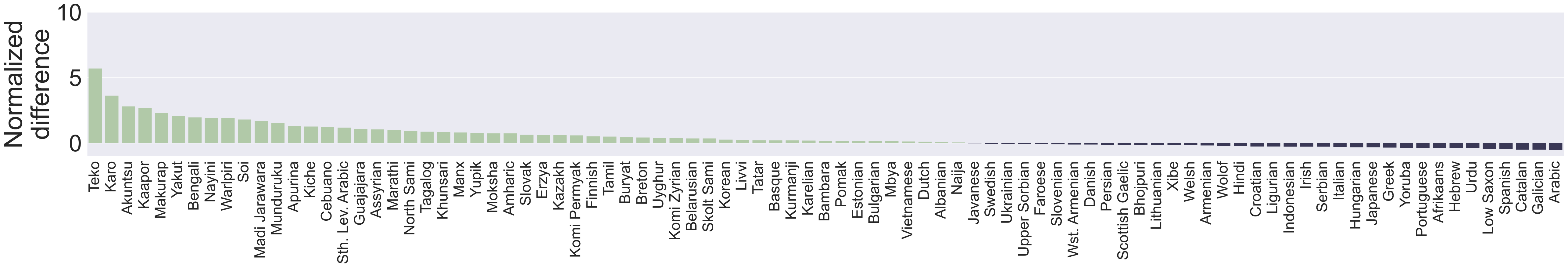}  
  \caption{Down sample English}
  \label{fig:sent_ds_en_tok}
\end{subfigure}
\begin{subfigure}{\textwidth}
  \centering
  \includegraphics[width=\textwidth]{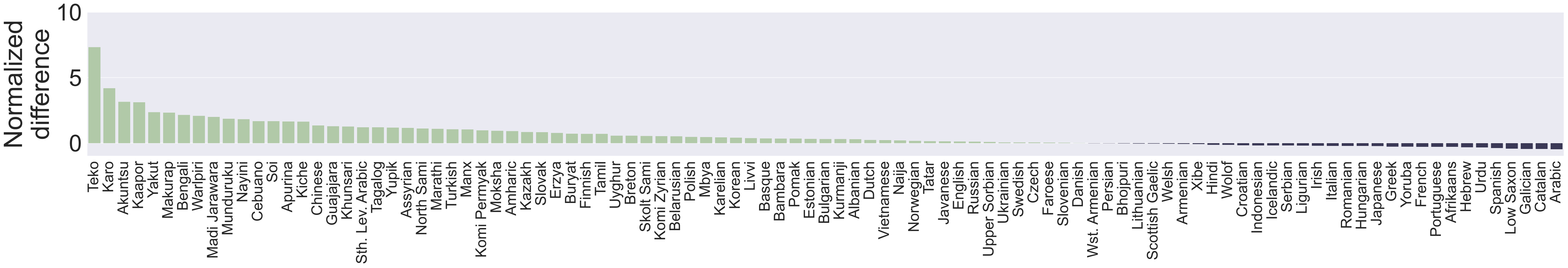}  
  \caption{Down sample German}
  \label{fig:sent_ds_de_tok}
\end{subfigure}
\begin{subfigure}{\textwidth}
  \centering
  \includegraphics[width=\textwidth]{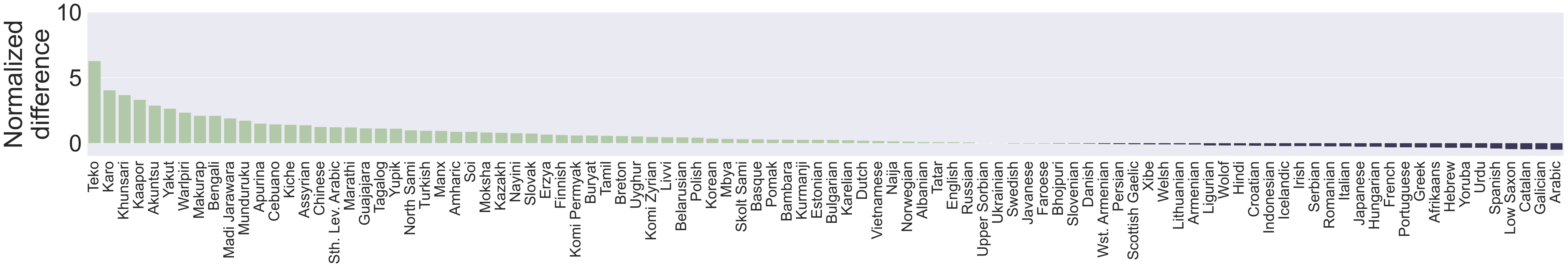}  
  \caption{Down sample Czech}
  \label{fig:sent_ds_cz_tok}
\end{subfigure}
\begin{subfigure}{\textwidth}
  \centering
  \includegraphics[width=\textwidth]{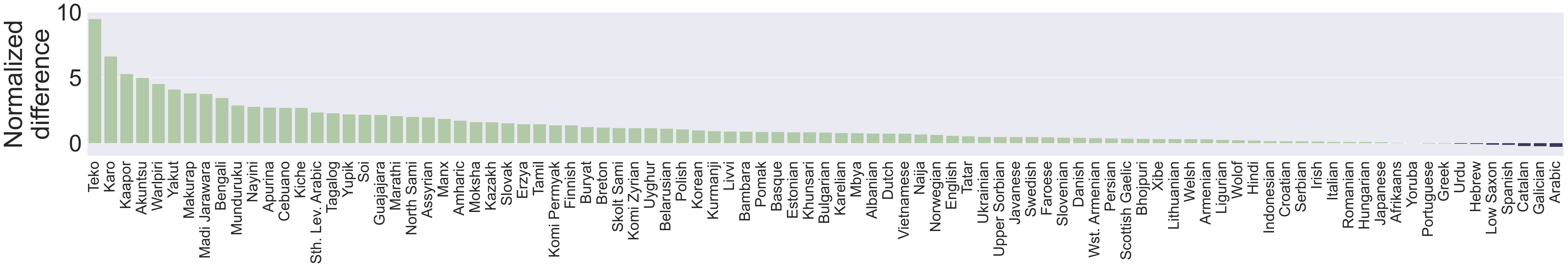}  
  \caption{Down sample French}
  \label{fig:sent_ds_fr_tok}
\end{subfigure}
\begin{subfigure}{\textwidth}
  \centering
  \includegraphics[width=\textwidth]{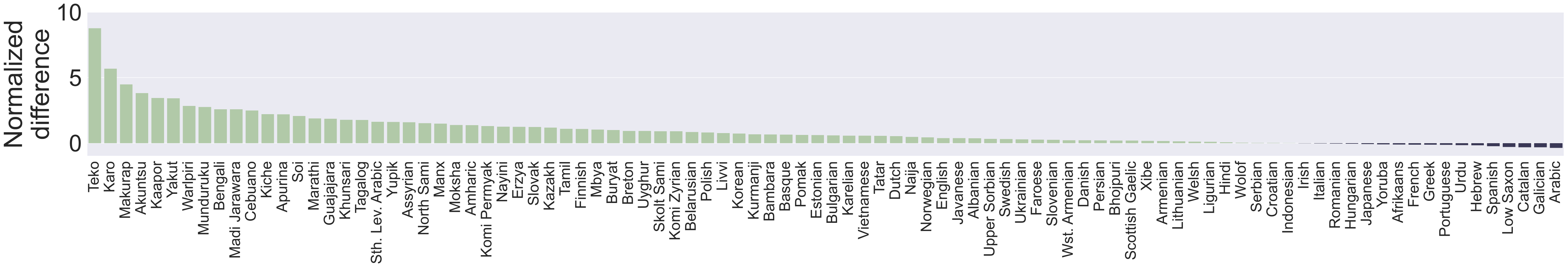}  
  \caption{Down sample Icelandic}
  \label{fig:sent_ds_ic_tok}
\end{subfigure}
\begin{subfigure}{\textwidth}
  \centering
  \includegraphics[width=\textwidth]{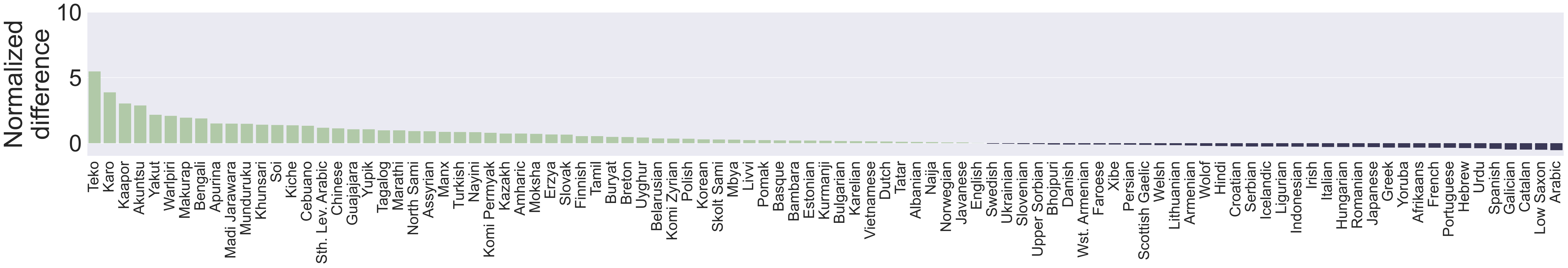}  
  \caption{Down sample Russian}
  \label{fig:sent_ds_ru_tok}
\end{subfigure}
\caption{Effect of down sampling on number of tokens. Down sampling based on number of sentences. Each plot describes a different language that we are down sampling. The x-axis describes the language that we use as reference.}
\label{fig:ds_effect_ud_stats_sent_based_toks_per_sents}
\end{figure*}

\subsection{Additional micro F1-Scores for POS-tagging Experiments}
\label{sec:appendix_pos_scores}

\begin{table*}
\centering
\begin{tabular}{rrrrrr}
\toprule
  & & & \multicolumn{3}{c}{\textbf{Micro F1}} \\
  \cmidrule(lr){4-6}
\multicolumn{1}{r}{\textbf{Vocab size}} & \multicolumn{1}{r}{\textbf{Nr Sents}} & \multicolumn{1}{r}{\textbf{Nr Toks}} & \multicolumn{1}{r}{\textbf{Word2Vec}} & \multicolumn{1}{r}{\textbf{Glove}} & \multicolumn{1}{r}{\textbf{BERT}} \\
\midrule
$1{,}000 $ & $1{,}000 $ & $7{,}235.75 \pm 174.485$ & $0.189 \pm 0.013$ & $0.581 \pm 0.021$ &  $0.801 \pm 0.007$ \\ 
$2{,}000 $ & $1{,}000 $ & $11{,}252.0 \pm 227.885$ & $0.208 \pm 0.015$ & $\textbf{0.624} \pm \textbf{0.017}$ &  $0.853 \pm 0.008$ \\ 
$3{,}000 $ & $1{,}000 $ & $14{,}867.2 \pm 292.534$ & $\textbf{0.215} \pm \textbf{0.005}$ & $0.622 \pm 0.030$ & $\textbf{0.880} \pm \textbf{0.015}$\\ 
\bottomrule
\end{tabular}
\caption{POS-tagging scores for different vocabulary sizes, and different word embeddings. Micro F1-scores.}
\label{tab:pos_tagging_micro_f1}
\end{table*}

\section{Additional Plots MT Experiments}
\label{sec:appendix_mt_plots}

\begin{figure*}
\begin{subfigure}{\textwidth}
  \centering
  \includegraphics[width=\textwidth]{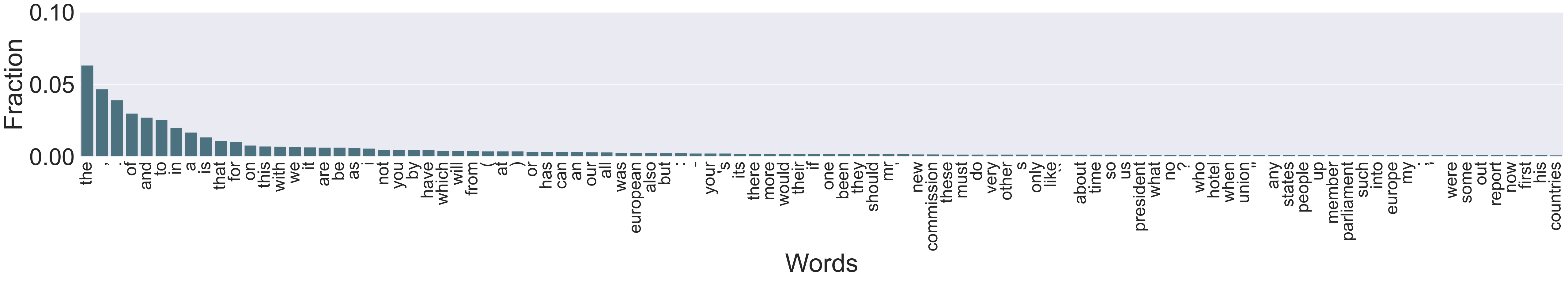}  
  \caption{WMT14}
  \label{fig:top_100_wmt14}
\end{subfigure}
\begin{subfigure}{\textwidth}
  \centering
  \includegraphics[width=\textwidth]{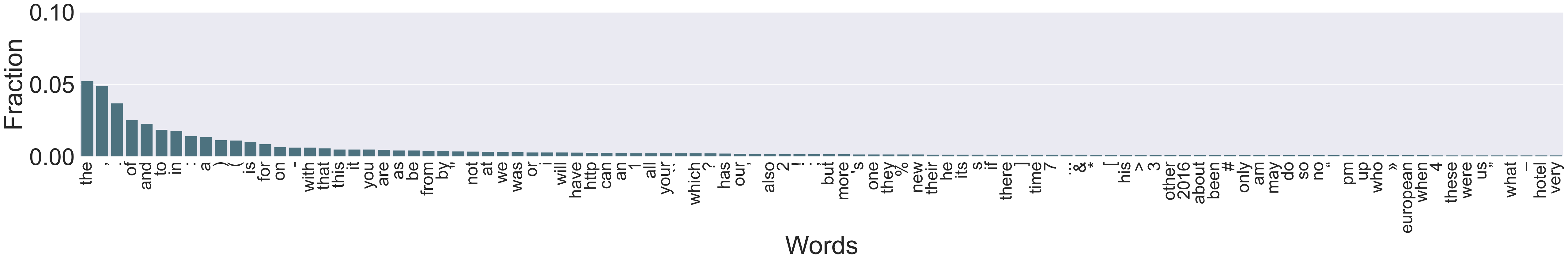}  
  \caption{WMT18}
  \label{fig:top_100_wmt8}
\end{subfigure}
\begin{subfigure}{\textwidth}
  \centering
  \includegraphics[width=\textwidth]{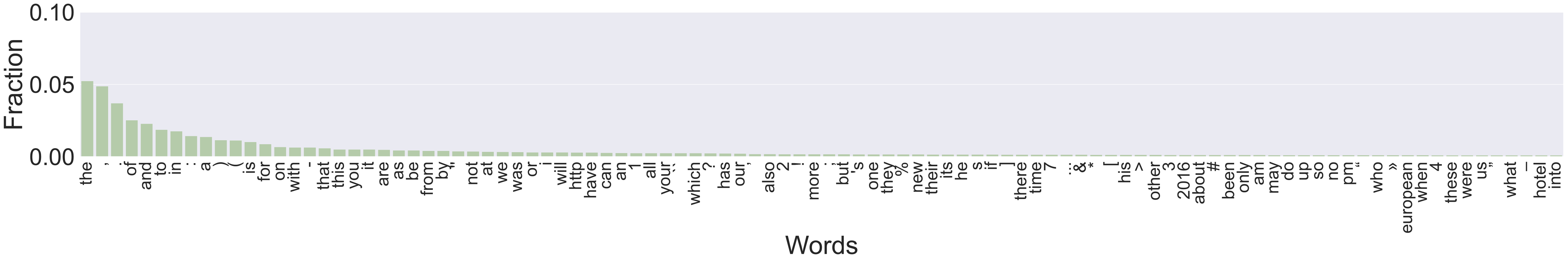}  
  \caption{Sentence based down sampled WMT18}
  \label{fig:top_100_wmt18_sent_ds}
\end{subfigure}
\begin{subfigure}{\textwidth}
  \centering
  \includegraphics[width=\textwidth]{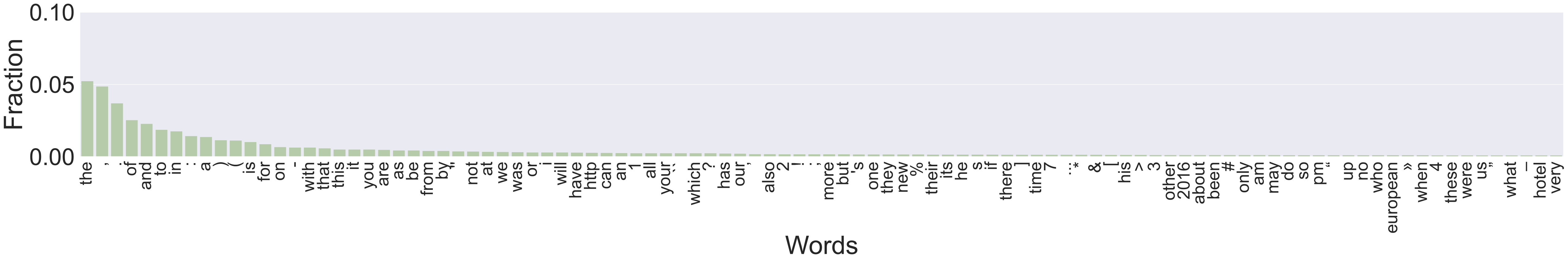}  
  \caption{Token based down sampled WMT18}
  \label{fig:top_100_wmt18_tok_ds}
\end{subfigure}
\caption{Top $100$ words in the train sets of WMT14, WMT18 and down sampled WMT18.}
\label{fig:wmt_vocab_dists}
\end{figure*}

\onecolumn 

\section{UD Languages}
\label{sec:appendix_ud_languages}

\begin{longtable}[c]{p{.22\textwidth} p{.35\textwidth}  p{.04\textwidth} p{.025\textwidth} p{.03\textwidth} p{.06\textwidth} p{.12\textwidth} } 
   \toprule
   \textbf{UD Language} & \textbf{UD Corpus} & \textbf{Train} & \textbf{Dev} & \textbf{Test} & \textbf{WALS} & \textbf{Wiki} \\
   \midrule
Afrikaans & UD AFRIKAANS & x & x & x &  & analytic \\
\midrule
Akuntsu & UD AKUNTSU &  &  & x &  &  \\
\midrule
Albanian & UD ALBANIAN &  &  & x &  &  \\
\midrule
Amharic & UD AMHARIC &  &  & x &  &  \\
\midrule
Apurina & UD APURINA &  &  & x & 6 &  \\
\midrule
Arabic & UD ARABIC-PADT & x & x & x & 6 &  \\
 & UD ARABIC-PUD &  &  & x &  &  \\
\midrule
Armenian & UD ARMENIAN-ArmTDP & x & x & x & 2 &  \\
 & UD ARMENIAN-BSUT & x & x & x &  &  \\
\midrule
Assyrian & UD ASSYRIAN &  &  & x &  &  \\
\midrule
Bambara & UD BAMBARA &  &  & x &  & agglutinative \\
\midrule
Basque & UD BASQUE & x & x & x & 4 & agglutinative \\
\midrule
Belarusian & UD BELARUSIAN & x & x & x &  &  \\
\midrule
Bengali & UD BENGALI &  &  & x &  &  \\
\midrule
Bhojpuri & UD BHOJPURI &  &  & x &  &  \\
\midrule
Breton & UD BRETON &  &  & x &  &  \\
\midrule
Bulgarian & UD BULGARIAN & x & x & x &  & analytic \\
\midrule
Buryat & UD BURYAT & x &  & x &  &  \\
\midrule
Catalan & UD CATALAN & x & x & x &  & agglutinative \\
\midrule
Cebuano & UD CEBUANO &  &  & x &  &  \\
\midrule
Chinese & UD CHINESE-GSD & x & x & x &  & analytic \\
 & UD CHINESE-GSDSimp & x & x & x &  &  \\
 & UD CLASSICAL CHINESE-Kyoto & x & x & x &  &  \\
\midrule
Croatian & UD CROATIAN & x & x & x &  &  \\
\midrule
Czech & UD CZECH-CAC & x & x & x &  & fusional \\
 & UD CZECH-CLTT & x & x & x &  &  \\
 & UD CZECH-FicTree & x & x & x &  &  \\
 & UD CZECH-PDT-l & x & x & x &  &  \\
 & UD CZECH-PDT-c & x &  &  &  &  \\
 & UD CZECH-PDT-m & x &  &  &  &  \\
 & UD CZECH-PDT-v & x &  &  &  &  \\
 & UD CZECH-PUD &  &  & x &  &  \\
\midrule
Danish & UD DANISH & x & x & x &  & analytic \\
\midrule
Dutch & UD DUTCH-Alpino & x & x & x &  &  \\
 & UD DUTCH-LassySmall & x & x & x &  &  \\
\midrule
English & UD ENGLISH-Atis & x & x & x & 2 & analytic \\
 & UD ENGLISH-EWT & x & x & x &  &  \\
 & UD ENGLISH-GUM & x & x & x &  &  \\
 & UD ENGLISH-LinES & x & x & x &  &  \\
 & UD ENGLISH-ParTUT & x & x & x &  &  \\
 & UD ENGLISH-Pronouns &  &  & x &  &  \\
 & UD ENGLISH-PUD &  &  & x &  &  \\
\midrule
Erzya & UD ERZYA &  &  & x &  & agglutinative \\
\midrule
Estonian & UD ESTONIAN-EDT & x & x & x &  & agglutinative \\
 & UD ESTONIAN-EWT & x & x & x &  &  \\
\midrule
Faroese & UD FAROESE-FarPaHC & x & x & x &  &  \\
 & UD FAROESE-OFT &  &  & x &  &  \\
\midrule
Finnish & UD FINNISH-FTB & x & x & x & 2 & agglutinative \\
 & UD FINNISH-OOD &  &  & x &  &  \\
 & UD FINNISH-PUD &  &  & x &  &  \\
 & UD FINNISH-TDT & x & x & x &  &  \\
\midrule
French & UD FRENCH-FQB &  &  & x & 4 &  \\
 & UD FRENCH-FTB & x & x & x &  &  \\
 & UD FRENCH-GSD & x & x & x &  &  \\
 & UD FRENCH-ParisStories & x &  & x &  &  \\
 & UD FRENCH-ParTUT & x & x & x &  &  \\
 & UD FRENCH-PUD &  &  & x &  &  \\
 & UD FRENCH-Rhapsodie & x & x & x &  &  \\
 & UD FRENCH-Sequoia & x & x & x &  &  \\
\midrule
Galician & UD GALICIAN-CTG & x & x & x &  &  \\
 & UD GALICIAN-TreeGal & x &  & x &  &  \\
\midrule
German & UD GERMAN-GSD & x & x & x & 2 &  \\
 & UD GERMAN-HDT-a1 & x & x & x &  &  \\
 & UD GERMAN-HDT-a2 & x &  &  &  &  \\
 & UD GERMAN-HDT-b1 & x &  &  &  &  \\
 & UD GERMAN-HDT-b2 & x &  &  &  &  \\
 & UD GERMAN-LIT &  &  & x &  &  \\
 & UD GERMAN-PUD &  &  & x &  &  \\
\midrule
Greek & UD GREEK & x & x & x & 4 &  \\
\midrule
Guajajara & UD GUAJAJARA &  &  & x &  & agglutinative \\
\midrule
Hebrew & UD HEBREW-IAHLTwiki & x & x & x & 4 &  \\
 & UD HEBREW-HTB & x & x & x &  &  \\
\midrule
Hindi & UD HINDI-HDTB & x & x & x & 2 &  \\
\midrule
Hungarian & UD HUNGARIAN & x & x & x & 4 & agglutinative \\
\midrule
Icelandic & UD ICELANDIC-IcePaHC & x & x & x &  &  \\
 & UD ICELANDIC-Modern & x & x & x &  &  \\
 & UD ICELANDIC-PUD &  &  & x &  &  \\
\midrule
Indonesian & UD INDONESIAN-CSUI & x &  & x & 4 &  \\
 & UD INDONESIAN-PUD &  &  & x &  &  \\
 & UD INDONESIAN-GSD & x & x & x &  &  \\
\midrule
Irish & UD IRISH-IDT & x & x & x &  &  \\
 & UD IRISH-TwittIrish &  &  & x &  &  \\
\midrule
Italian & UD ITALIAN-ISDT & x & x & x &  &  \\
 & UD ITALIAN-MarkIT & x & x & x &  &  \\
 & UD ITALIAN-ParTUT & x & x & x &  &  \\
 & UD ITALIAN-PoSTWITA & x & x & x &  &  \\
 & UD ITALIAN-PUD &  &  & x &  &  \\
 & UD ITALIAN-TWITTIRO & x & x & x &  &  \\
 & UD ITALIAN-Valico &  &  & x &  &  \\
 & UD ITALIAN-VIT & x & x & x &  &  \\
\midrule
Japanese & UD JAPANESE-GSD & x & x & x & 4 & agglutinative \\
 & UD JAPANESE-Modern &  &  & x &  &  \\
 & UD JAPANESE-PUD &  &  & x &  &  \\
\midrule
Javanese & UD JAVANESE &  &  & x &  & agglutinative \\
\midrule
Kaapor & UD KAAPOR &  &  & x &  &  \\
\midrule
Karelian & UD KARELIAN &  &  & x &  &  \\
\midrule
Karo & UD KARO &  &  & x &  &  \\
\midrule
Kazakh & UD KAZAKH & x &  & x &  & agglutinative \\
\midrule
Khunsari & UD KHUNSARI &  &  & x &  &  \\
\midrule
Kiche & UD KICHE &  &  & x &  &  \\
\midrule
Komi permyak & UD KOMI PERMYAK &  &  & x &  & agglutinative \\
\midrule
Komi zyrian & UD KOMI ZYRIAN-IKDP &  &  & x &  &  \\
 & UD KOMI ZYRIAN-Lattice &  &  & x &  &  \\
\midrule
Korean & UD KOREAN-GSD & x & x & x & 6 & agglutinative \\
 & UD KOREAN-Kaist & x & x & x &  &  \\
 & UD KOREAN-PUD &  &  & x &  &  \\
\midrule
Kurmanji & UD KURMANJI & x &  & x &  &  \\
\midrule
Ligurian & UD LIGURIAN & x &  & x &  &  \\
\midrule
Lithuanian & UD LITHUANIAN-ALKSNIS & x & x & x &  &  \\
 & UD LITHUANIAN-HSE & x & x & x &  &  \\
\midrule
Livvi & UD LIVVI & x &  & x &  &  \\
\midrule
Low saxon & UD LOW SAXON &  &  & x &  &  \\
\midrule
Madi jarawara & UD MADI JARAWARA &  &  & x &  & agglutinative \\
\midrule
Makurap & UD MAKURAP &  &  & x &  &  \\
\midrule
Manx & UD MANX &  &  & x &  &  \\
\midrule
Marathi & UD MARATHI & x & x & x &  &  \\
\midrule
Mbya & UD MBYA GUARANI-Thomas &  &  & x &  &  \\
\midrule
Moksha & UD MOKSHA &  &  & x &  & agglutinative \\
\midrule
Munduruku & UD MUNDURUKU &  &  & x &  &  \\
\midrule
Naija & UD NAIJA & x & x & x &  &  \\
\midrule
Nayini & UD NAYINI &  &  & x &  &  \\
\midrule
North sami & UD NORTH SAMI & x &  & x &  & agglutinative \\
\midrule
Norwegian & UD NORWEGIAN Bokmaal & x & x & x &  & analytic \\
 & UD NORWEGIAN Nynorsk & x & x & x &  &  \\
 & UD NORWEGIAN NynorskLIA & x & x & x &  &  \\
\midrule
Persian & UD PERSIAN-PerDT & x & x & x & 4 &  \\
 & UD PERSIAN-Seraji & x & x & x &  &  \\
\midrule
Polish & UD POLISH-LFG & x & x & x &  & fusional \\
 & UD POLISH-PDB & x & x & x &  &  \\
 & UD POLISH-PUD &  &  & x &  &  \\
\midrule
Pomak & UD POMAK & x & x & x &  &  \\
\midrule
Portuguese & UD PORTUGUESE-BOSQUE & x & x & x &  &  \\
 & UD PORTUGUESE-GSD & x & x & x &  &  \\
 & UD PORTUGUESE-PUD &  &  & x &  &  \\
\midrule
Romanian & UD Romanian-ArT &  &  & x &  &  \\
 & UD Romanian-Nonstandard & x & x & x &  &  \\
 & UD Romanian-RRT & x & x & x &  &  \\
 & UD Romanian-SiMoNERo & x & x & x &  &  \\
\midrule
Russian & UD RUSSIAN-GSD & x & x & x & 4 & fusional \\
 & UD RUSSIAN-PUD &  &  & x &  &  \\
 & UD RUSSIAN-SynTagRus-a & x & x & x &  &  \\
 & UD RUSSIAN-SynTagRus-b & x &  &  &  &  \\
 & UD RUSSIAN-SynTagRus-c & x &  &  &  &  \\
 & UD RUSSIAN-Taiga & x & x & x &  &  \\
\midrule
Scottish gaelic & UD SCOTTISH GAELIC & x & x & x &  &  \\
\midrule
Serbian & UD SERBIAN & x & x & x &  &  \\
\midrule
Skolt sami & UD SKOLT SAMI &  &  & x &  & fusional \\
\midrule
Slovak & UD SLOVAK & x & x & x &  &  \\
\midrule
Slovenian & UD SLOVENIAN-SSJ & x & x & x &  &  \\
 & UD SLOVENIAN-SST & x &  & x &  &  \\
\midrule
Soi & UD SOI &  &  & x &  &  \\
\midrule
South levantine arabic & UD SOUTH LEVANTINE ARABIC &  &  & x &  &  \\
\midrule
Spanish & UD SPANISH-AnCora & x & x & x & 4 &  \\
 & UD SPANISH-GSD & x & x & x &  &  \\
 & UD SPANISH-PUD &  &  & x &  &  \\
\midrule
Swedish & UD SWEDISH LinES & x & x & x &  & analytic \\
\midrule
Tagalog & UD TAGALOG-TRG &  &  & x & 2 &  \\
 & UD TAGALOG-Ugnayan &  &  & x &  &  \\
\midrule
Tamil & UD Tamil-MWTT &  &  & x &  & agglutinative \\
 & UD Tamil-TTB & x & x & x &  &  \\
\midrule
Tatar & UD TATAR &  &  & x &  & agglutinative \\
\midrule
Teko & UD TEKO &  &  & x &  &  \\
\midrule
Turkish & UD TURKISH-Atis & x & x & x & 6 & agglutinative \\
 & UD TURKISH-BOUN & x & x & x &  &  \\
 & UD TURKISH-FrameNet & x & x & x &  &  \\
 & UD TURKISH-GB &  &  & x &  &  \\
 & UD TURKISH-IMST & x & x & x &  &  \\
 & UD TURKISH-Kenet & x & x & x &  &  \\
 & UD TURKISH-Penn & x & x & x &  &  \\
 & UD TURKISH-PUD &  &  & x &  &  \\
 & UD TURKISH-Tourism & x & x & x &  &  \\
\midrule
Ukrainian & UD UKRAINIAN & x & x & x &  &  \\
\midrule
Upper sorbian & UD UPPER SORBIAN & x &  & x &  &  \\
\midrule
Urdu & UD URDU & x & x & x &  &  \\
\midrule
Uyghur & UD UYGHUR & x & x & x &  & agglutinative \\
\midrule
Vietnamese & UD VIETNAMESE & x & x & x & 0 & analytic \\
\midrule
Warlpiri & UD WARLPIRI &  &  & x &  &  \\
\midrule
Welsh & UD WELSH & x & x & x &  &  \\
\midrule
Western armenian & UD WESTERN ARMENIAN & x & x & x &  &  \\
\midrule
Wolof & UD WOLOF & x & x & x &  &  \\
\midrule
Xibe & UD XIBE &  &  & x &  &  \\
\midrule
Yakut & UD YAKUT &  &  & x &  & agglutinative \\
\midrule
Yoruba & UD YORUBA &  &  & x & 6 & analytic \\
\midrule
Yupik & UD YUPIK &  &  & x &  & agglutinative \\

     \bottomrule
    \caption{Languages and corpora from the UD included in the POS-tagging experiments.} 
   \label{tab:ud_languages}
\end{longtable}

\end{document}